\pdfoutput=1
\documentclass[11pt]{article}

\usepackage[final]{acl}

\usepackage{times}
\usepackage{latexsym}
\usepackage{enumitem}
\usepackage[T1]{fontenc}
\usepackage[utf8]{inputenc}

\usepackage{microtype}

\usepackage{inconsolata}

\usepackage{graphicx}

\usepackage{subcaption}
\usepackage{multirow}
\usepackage{booktabs}
\usepackage{adjustbox}
\definecolor{green}{rgb}{0.05, 0.9, 0.05}

\definecolor{redbw}{HTML}{d7191c}
\definecolor{greenbw}{HTML}{1c8036}
\definecolor{bluebw}{HTML}{2b83ba}

\newcommand\mup[1]{\footnotesize{\color{greenbw}#1\color{black}}}
\newcommand\mdo[1]{\footnotesize{\color{redbw}#1\color{black}}}
\newcommand\msa[1]{\footnotesize{#1}}

\newcommand{\varcell}[1]{\scriptsize$\pm${#1}}

\usepackage{amssymb}
\usepackage{amsmath}
\usepackage{enumitem}
\usepackage{cleveref}
\usepackage{soul}
\usepackage{float}
\usepackage{appendix}
\usepackage{tabularx}
\usepackage{listings}
\usepackage{lscape}

\title{EmoGist: Efficient In-Context Learning\\for Visual Emotion Understanding}

\author{Ronald Seoh \and Dan Goldwasser \\
         Purdue University \\ \texttt{\{bseoh, dgoldwas\}@purdue.edu}}

\begin{document}

\maketitle

\begin{abstract}
In this paper, we introduce EmoGist, a training-free, in-context learning method for performing visual emotion classification with LVLMs. The key intuition of our approach is that context-dependent definition of emotion labels could allow more accurate predictions of emotions, as the ways in which emotions manifest within images are highly context dependent and nuanced. EmoGist pre-generates multiple descriptions of emotion labels, by analyzing the clusters of example images belonging to each label. At test time, we retrieve a version of description based on the cosine similarity of test image to cluster centroids, and feed it together with the test image to a fast LVLM for classification. Through our experiments, we show that EmoGist allows up to 12 points improvement in micro F1 scores with the multi-label Memotion dataset, and up to 8 points in macro F1 in the multi-class FI dataset.
\end{abstract}

\section{Introduction}

\begin{figure}[t]
\centering
\includegraphics[width=\columnwidth]{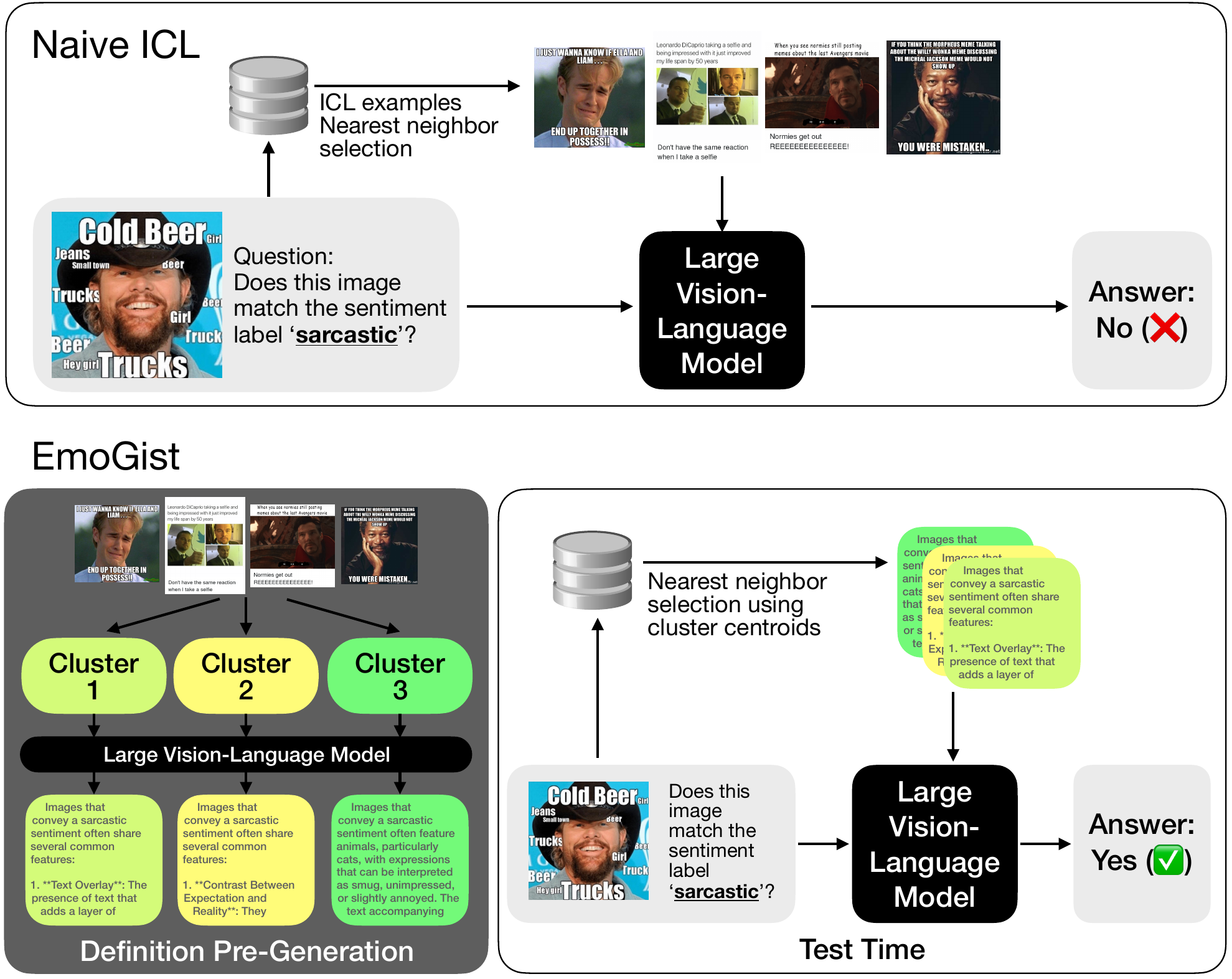}
\caption{For visual emotion classification, naive in-context learning (ICL) struggles as providing multiple nuanced visual examples often lead LVLMs to make incorrect predictions. EmoGist guides LVLMs using pre-generated multiple descriptions of emotion labels obtained by analyzing the clusters of example images.}
\label{fig:overview}
\end{figure}

Automated classification of visual emotion \citep{ekman1993facial,lang1999international,mikels2005emotional} is an extremely challenging problem, as the ways in which emotions are embedded within images are inherently nuanced. Hence, even large vision-language models (LVLMs) that are extensively trained for reasoning over visual inputs struggle in detecting these emotions \citep{bhattacharyya-wang-2025-evaluating}, as their training may not necessarily involve the ability to understand such nuanced patterns.

In this paper, we introduce EmoGist, a training-free, in-context learning method for performing visual emotion classification with LVLMs. The key intuition of our approach is that the real meaning of different emotion labels could be dependent on the image's context. For example, we could intuitively imagine that the way the emotion of `excitement' for sporting events could be significantly different from the `excitement' of the academics for an upcoming conference. Hence, guiding LVLMs with such context dependent definition of emotion labels could allow the models to better focus on the nuanced patterns of the image.

EmoGist automatically pre-generates nuanced, context-specific descriptions of emotion labels, by analyzing the clusters of example images belonging to each label. At test time, we retrieve a version of description based on the cosine similarity of test image to cluster centroids, and feed it together with the test image to a fast LVLM for classification. Through our experiments, we show that EmoGist allows up to 12 points improvement in micro F1 scores with the multi-label emotion classification, and up to 8 points improvement in macro F1 for the multi-class case. We also demonstrate that EmoGist could achieve improvements in smaller LVLMs with 2 billion parameters.\footnote{All the program codes used to produce results presented in this paper are available at \url{https://tinyurl.com/emo-gist}.}

\section{Related Work}

While visual emotions has been extensively studied in the many related fields including computer vision and psychology \citep{mikels2005emotional,10.1145/1873951.1873965,7298687,You_Luo_Jin_Yang_2016,10376578}, we are only starting to see the efforts to exploit large vision-language models for automated understanding of visual emotions \citep{Xie_2024_CVPR,10802538,xenos2024vllmsprovidebettercontext,pmlr-v260-lei25a,bhattacharyya-wang-2025-evaluating}.

Visual in-context learning (ICL) has seen considerable amount of interest in recent literature \citep{NEURIPS2023_398ae57e,zhou-etal-2024-visual,Zhang_2024_WACV}, where many work have investigated effective strategies for choosing visual ICL examples given the test instance. However, we believe that our work is first to investigate ICL strategies with LVLMs in detail for evoked emotion classification.

\section{EmoGist}
\label{sec:emogist}

We describe the major components of EmoGist, our in-context learning method with LVLMs for emotion classification. Instead of naively retrieving individual examples based on the visual similarity of the image, the key idea is to obtain an nuanced description of emotion labels, which could effectively serve as the decision boundary for the LVLM to make its predictions on.

Because the same emotion could manifest in many different ways for across different images, we develop a strategy where we utilize stronger LVLMs to pre-generate multiple descriptions of different emotion labels, by analyzing the clusters of example images belonging to each category.

\begin{figure*}[t!]
\centering
\includegraphics[width=\linewidth]{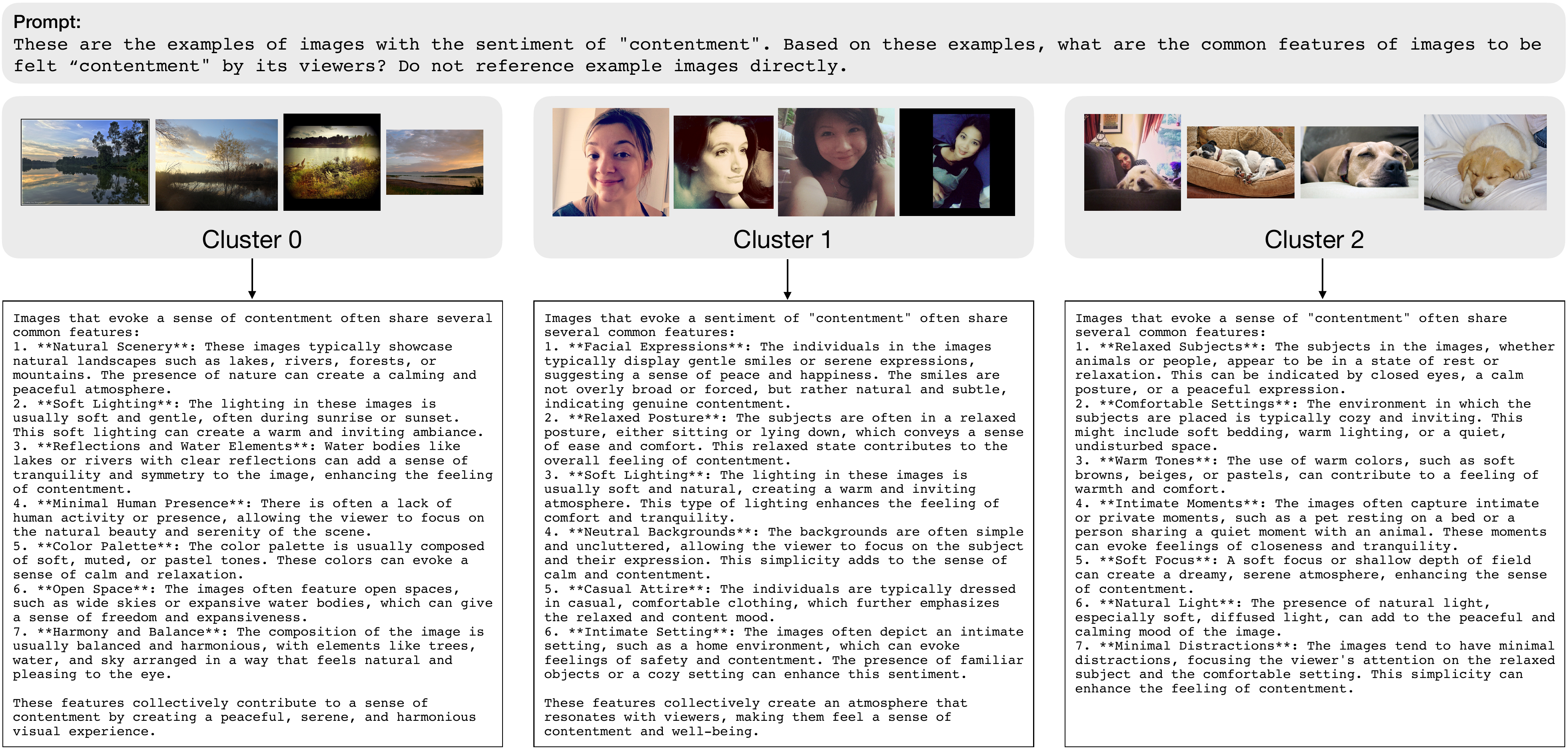}
\caption{Example label descriptions for clusters within the \texttt{contentment} class of the FI dataset, as generated by the Qwen2.5-VL 72B model.}
\label{fig:label_desc_examples}
\end{figure*}

\paragraph{Embedding and storing the pool of emotion label examples} In order to generate multiple descriptions of emotion labels, we begin by embedding the pool of example images with an embedding model. We use the MM-E5 model, the state-of-the-art multimodal embedding model by \citet{chen2025mme5improvingmultimodalmultilingual}. Then we store the embeddings into a \textsc{hnswlib} vector database \citep{malkov2018efficient}.

\paragraph{Clustering} After creating the vector database of example images, we run the $k$-means clustering algorithm \citep{1056489} against the set of embeddings to get different clusters. Because we are interested in creating multiple versions of descriptions for specific labels, clustering is done separately for each emotion label. We tune the hyperparameter $k$ by setting a portion of example images aside as a validation set, and tune by evaluating the end task performance on them. Due to the limited computational resources available, we only experiment with the $k$ values of 2, 4, and 6.\footnote{Please see \autoref{sec:hyperparameter_tuning} for hyperparameter tuning procedures and sensitivity analysis.}

\paragraph{Generating label descriptions} With the cluster information, we provide a strong LVLM with images from each cluster, and prompt them to explain why the given images belong to the emotion label. For our experiments, we use the Qwen2.5-VL 72B model \citep{bai2025qwen25vltechnicalreport} for generating descriptions. As it is not possible to provide the LVLM with all images from the cluster due to GPU memory limits and context length, we select 4 images from each cluster to create one version of label description. \autoref{fig:label_desc_examples} shows the prompt used and examples of generated descriptions.

\paragraph{Selection of clusters at test time} We note that EmoGist addresses both multi-label and multi-class classification cases. For multi-label classification, we assume that all test instances are binarized, and perform predictions for each candidate label given the image. For each candidate label, we retrieve the closest cluster among the candidates with the corresponding label.

For multi-class classification where the classes are exclusive to each other, we perform the classification only once by providing the list of all candidate classes to the model. We perform the search across the entire set of clusters regardless of their classes, and use the closest cluster to the test image in terms of its distance to the centroid.

Once the cluster and associated label description has been chosen, we prepend the description to the test image and classification prompt.\footnote{Please see \autoref{sec:prompt_classification} for the full prompts used for classification.}

\paragraph{Ensembles} As we only select a subset of images from each cluster for generating an description, it may be the case that the selected images and descriptions may not sufficiently match the test image. In order to mitigate this issue, we introduce a simple ensemble scheme, where we generate multiple versions of descriptions for each cluster, perform multiple predictions against a single example and take the majority vote. We generate multiple descriptions for the cluster by ranking all images within the cluster by their distance to the centroid, and generate descriptions for every top 4 images in the ranked list.

\section{Experiments}

\begin{table*}[ht]
    \centering
    \begin{subtable}[h!]{0.8\textwidth}
    \begin{adjustbox}{width=\textwidth}
    \begin{tabular}{cccc|ccc|ccc}
        %\toprule
        \multirow{3}{*}{\textbf{Method}} & \multicolumn{9}{c}{\textbf{Model}} \\
                                         & \multicolumn{3}{c}{Qwen2.5-VL 7B} & \multicolumn{3}{c}{Aya Vision 8B} & \multicolumn{3}{c}{InternVL2.5 8B-MPO}  \\
                                         & Precision & Recall & F1     & Precision & Recall & F1     & Precision & Recall & F1 \\
        \midrule
        \multirow{1.5}{*}{\centering Zero-Shot} & 47.121 \vspace{-0.5em} & 50.381    & 40.089    & 50.094   & 47.181 & 44.188  & 47.237 & 50.379 & 42.939 \\
                                                & \varcell{0.065} \vspace{-0.5em} & \varcell{0.090} & \varcell{0.065} & \varcell{0.285} & \varcell{0.188} & \varcell{0.102} & \varcell{0.297} & \varcell{0.171} & \varcell{0.223} \\
        \specialrule{.2em}{.3em}{.3em}
        \multirow{2.25}{*}{\centering Global Exp}            & 30.503 \vspace{-0.5em} & 30.156 & 23.480 & 26.416 & 20.782 & 15.405 & 32.242 & 35.224 & 30.583 \\
                                                          & \varcell{0.325} \vspace{-0.5em} & \varcell{0.439} & \varcell{0.404} & \varcell{1.670} & \varcell{0.468} & \varcell{0.322} & \varcell{0.316} & \varcell{0.452} & \varcell{0.312} \\
                                                          & \mdo{-16.619} & \mdo{-20.226} & \mdo{-16.609} & \mdo{-23.678} & \mdo{-26.398} & \mdo{-28.783} & \mdo{-14.996} & \mdo{-15.155} & \mdo{-12.355} \\
        \midrule
        \multirow{2.25}{*}{\centering $\text{ICL}_{\text{sim}}$} & 48.656 \vspace{-0.5em} & 49.898 & 42.733 & 48.649 & 42.444 & 35.257 & 42.613 & 43.658 & 37.653 \\
                                                        & \varcell{0.029} \vspace{-0.5em} & \varcell{0.022} & \varcell{0.016} & \varcell{0.592} & \varcell{0.122} & \varcell{0.185} & \varcell{0.215} & \varcell{0.171} & \varcell{0.161} \\
                                                          & \mup{+1.534} & \mdo{-0.483} & \mup{+2.643} & \mdo{-1.445} & \mdo{-4.736} & \mdo{-8.931} & \mdo{-4.624} & \mdo{-6.721} & \mdo{-5.286} \\
        \midrule
        \multirow{2.25}{*}{\centering $\text{ICL}_{\text{all}}$} & 23.464 \vspace{-0.5em} & 14.633 & 5.664 & 13.223 & 12.609 & 1.511 & 16.682 & 16.107 & 13.454 \\
                                                        & \varcell{0.948} \vspace{-0.5em} & \varcell{0.024} & \varcell{0.081} & \varcell{3.242} & \varcell{0.042} & \varcell{0.102} & \varcell{0.517} & \varcell{0.259} & \varcell{0.298} \\
                                                          & \mdo{-23.657} & \mdo{-35.749} & \mdo{-34.425} & \mdo{-36.871} & \mdo{-34.572} & \mdo{-42.677} & \mdo{-30.555} & \mdo{-34.272} & \mdo{-29.485} \\
        \specialrule{.2em}{.3em}{.3em}
        \multirow{2.25}{*}{\centering $\text{EmoGist}_{\text{n}}$} & \textbf{52.944} \vspace{-0.5em} & \textbf{52.163} & \textbf{48.497} & 52.579 & \textbf{51.389} & \textbf{47.906} & \textbf{52.592} & \textbf{51.767} & \textbf{48.094} \\
                                                        & \varcell{0.364} \vspace{-0.5em} & \varcell{0.265} & \varcell{0.572} & \varcell{0.381} & \varcell{0.271} & \varcell{0.565} & \varcell{0.261} & \varcell{0.220} & \varcell{0.303} \\
                                                          & \mup{+5.822} & \mup{+1.782} & \mup{+8.408} & \mup{+2.485} & \mup{+4.208} & \mup{+3.718} & \mup{+5.354} & \mup{+1.388} & \mup{+5.155} \\
        \midrule
        \multirow{2.25}{*}{\centering $\text{EmoGist}_{\text{e}}$} & 52.772 \vspace{-0.5em} & 51.704 & 48.118 & \textbf{52.632} & 51.146 & 47.579 & 52.449 & 51.712 & 47.795 \\
                                                        & \varcell{0.377} \vspace{-0.5em} & \varcell{0.235} & \varcell{0.552} & \varcell{0.371} & \varcell{0.316} & \varcell{0.599} & \varcell{0.158} & \varcell{0.202} & \varcell{0.167} \\
                                                          & \mup{+5.650} & \mup{+1.323} & \mup{+8.029} & \mup{+2.538} & \mup{+3.965} & \mup{+3.391} & \mup{+5.211} & \mup{+1.333} & \mup{+4.856} \\
        \bottomrule
    \end{tabular}
    \end{adjustbox}
    \caption{FI}
    \end{subtable}
    %\hspace{100em}
    \begin{subtable}[h!]{0.8\textwidth}
    \begin{adjustbox}{width=\textwidth}
    \begin{tabular}{cccc|ccc|ccc}
        \multirow{1.5}{*}{\centering Zero-Shot} & 77.343 \vspace{-0.5em} & 48.260 & 59.434 & 75.799    & 63.814 & 69.292 & 73.520 & 63.377 & 68.070 \\
                                            & \varcell{0.013} \vspace{-0.5em} & \varcell{0.104} & \varcell{0.080} & \varcell{0.023} & \varcell{0.065} & \varcell{0.035} & \varcell{0.195} & \varcell{0.358} & \varcell{0.254} \\
        \specialrule{.2em}{.3em}{.3em}
        \multirow{2.25}{*}{\centering Global Exp}  & 77.134 \vspace{-0.5em} & 35.206 & 48.159 & 73.430 & 68.159 & 70.665 & 74.885 & 62.160 & 67.928 \\
                                                   & \varcell{0.180} \vspace{-0.5em} & \varcell{2.182} & \varcell{1.976} & \varcell{0.702} & \varcell{0.654} & \varcell{0.116} & \varcell{0.191} & \varcell{0.360} & \varcell{0.234} \\
                                                   & \mdo{-0.209} & \mdo{-13.054} & \mdo{-11.276} & \mdo{-2.369} & \mup{+4.345} & \mup{+1.373} & \mup{+1.365} & \mdo{-1.217} & \mdo{-0.142} \\
        \midrule
        \multirow{2.25}{*}{\centering $\text{ICL}_{\text{sim}}$} & 75.061 \vspace{-0.5em} & 64.642 & 69.462 & \textbf{76.721} & 60.915 & 67.910 & 72.601 & 55.476 & 62.890 \\
                                                   & \varcell{0.025} \vspace{-0.5em} & \varcell{0.107} & \varcell{0.071} & \varcell{0.043} & \varcell{0.077} & \varcell{0.062} & \varcell{0.101} & \varcell{0.432} & \varcell{0.290} \\
                                                   & \mdo{-2.282} & \mup{+16.381} & \mup{+10.028} & \mup{+0.922} & \mdo{-2.900} & \mdo{-1.382} & \mdo{-0.918} & \mdo{-7.901} & \mdo{-5.181} \\
        \specialrule{.2em}{.3em}{.3em}
        \multirow{2.25}{*}{\centering $\text{EmoGist}_{\text{n}}$} & 75.610 \vspace{-0.5em} & 62.693 & 68.540 & 68.693 & 84.265 & 75.611 & 72.162 & 72.438 & 72.289 \\
                                                   & \varcell{0.183} \vspace{-0.5em} & \varcell{0.492} & \varcell{0.237} & \varcell{0.548} & \varcell{1.958} & \varcell{0.747} & \varcell{0.204} & \varcell{0.708} & \varcell{0.338} \\
                                                   & \mdo{-1.732} & \mup{+14.432} & \mup{+9.105} & \mdo{-7.107} & \mup{+20.451} & \mup{+6.319} & \mdo{-1.358} & \mup{+9.061} & \mup{+4.219} \\
        \midrule
        \multirow{2.25}{*}{\centering $\text{EmoGist}_{\text{e}}$} & \textbf{78.682} \vspace{-0.5em} & \textbf{65.374} & \textbf{71.411} & 70.263 & \textbf{87.165} & \textbf{77.795} & \textbf{75.898} & \textbf{74.596} & \textbf{75.240} \\
                                                   & \varcell{0.067} \vspace{-0.5em} & \varcell{0.315} & \varcell{0.188} & \varcell{0.314} & \varcell{0.647} & \varcell{0.171} & \varcell{0.257} & \varcell{0.173} & \varcell{0.170} \\
                                                   & \mup{+1.339} & \mup{+17.114} & \mup{+11.977} & \mdo{-5.536} & \mup{+23.350} & \mup{+8.502} & \mup{+2.379} & \mup{+11.219} & \mup{+7.170} \\
        \bottomrule
    \end{tabular}
    \end{adjustbox}
    \caption{Memotion}
    \end{subtable}
    \caption{Results of our methods and baselines. We report macro scores for FI and micro scores for Memotion. All scores are averaged over six random seeds. We show standard errors as their confidence intervals. Boldfaces indicate the best performance for each metric across all methods for each model. Green and red numbers indicate the performance changes over the zero-shot baseline.}
    \label{tab:results}
\end{table*}

We used two datasets: one is the Memotion 1.0 dataset \citep{sharma-etal-2020-semeval,jin-etal-2024-mm}, a collection of meme images from social media where we perform a multi-label classification across 4 labels: \texttt{sarcastic}, \texttt{humorous}, \texttt{offensive}, and \texttt{motivational}. The second dataset is the FI dataset \citep{You_Luo_Jin_Yang_2016}, a collection of everyday images across the internet tagged based on the Ekman model \citep{ekman1993facial} of 8 classes: \texttt{amusement}, \texttt{anger}, \texttt{awe}, \texttt{contentment}, \texttt{disgust}, \texttt{excitement}, \texttt{fear}, and \texttt{sadness}.\footnote{Please see \autoref{sec:dataset} for more detailed dataset statistics.}

\subsection{Baselines}

In order to save computational resources and make our discussions more clear, we choose three different SOTA LVLMs of similar sizes for our experiments: Qwen2.5-VL 7B \citep{bai2025qwen25vltechnicalreport}, Aya Vision 8B \citep{dash2025ayavisionadvancingfrontier}, and InternVL2.5 8B-MPO \citep{wang2024mpo}. To ensure that our findings are not tied to particular pretraining, these three VLMs were chosen to ensure that they do not share the same image encoder or backbone LLM. We additionally provide a subset of results for smaller LVLMs in \autoref{tab:smol}.

We compare EmoGist with the following comparable in-context learning methods: 
    \begin{itemize}[noitemsep]
    \item Zero-Shot: We simply prompt a LVLM with the test image and prediction prompt.
    \item Global Exp: Instead of providing example images for a description, we prompt a large LVLM for ``global" description, where we ask the model to describe the common features of the images with the candidate emotion label, without providing any references.
    \item $\text{ICL}_{\text{sim}}$: We retrieve 4 images from the pool of images based on cosine similarity, regardless of their labels. This is closest to $\text{EmoGist}_{\text{n}}$, in terms of the number of examples.
    \item $\text{ICL}_{\text{all}}$: We also test another case of performing ICL for multiclass classification, where we provide one image each for all classes. Note that the FI dataset have 8 classes, resulting in 8 example images to be provided to a LVLM.
\end{itemize}

We also test two variants of EmoGist:
\begin{itemize}[noitemsep]
\item $\text{EmoGist}_{\text{n}}$: This variant of EmoGist uses 4 images from the cluster for description generation.
\item $\text{EmoGist}_{\text{e}}$: This variant of EmoGist performs ensembling, where we generate 3 versions of label description, each of them using 4 images without any overlap between them.
\end{itemize}

\subsection{Results}

\textbf{EmoGist achieves robust performance gains over all the baselines.} In \autoref{tab:results}, we can see that both $\text{EmoGist}_{\text{n}}$ and $\text{EmoGist}_{\text{e}}$ achieve consistent improvements over all the baselines. For FI, 
$\text{EmoGist}_{\text{n}}$ gains 8.41 points in terms of macro F1 score over Zero-Shot, and 5.76 points over $\text{ICL}_{\text{sim}}$. The trend is largely similar for Memotion, where $\text{EmoGist}_{\text{e}}$ gains 11.98 points in terms of micro F1 score over Zero-Shot, and 1.95 points over $\text{ICL}_{\text{sim}}$.

It is interesting to note that Global Exp, which is essentially providing the strong LVLM's general knowledge about emotion labels, is considerably worse than the Zero-Shot baseline. Therefore, we could see that having EmoGist's localized, cluster-specific label description makes a substantial difference. Lastly, adding ensembling $\text{EmoGist}_{\text{e}}$ shows consistent improvements over $\text{EmoGist}_{\text{n}}$, with notable gains in precision for Memotion.

\textbf{Naively performing visual ICL could be detrimental.} Another observation from \autoref{tab:results} is that our ICL baselines, $\text{ICL}_{\text{sim}}$ and $\text{ICL}_{\text{all}}$, are either under-performing, or marginally better than the Zero-Shot baselines. In the most extreme case, $\text{ICL}_{\text{all}}$ for FI sees over 42 points drop in F1 score, way below random guessing. In addition, while ICL achieves considerable performance with Qwen2.5-VL 7B on Memotion, Aya Vision 8B and InternVL 2.5 8B-MPO failed to achieve comparable scores. As reasoning over multiple image inputs is still an area of active research in pretraining and post-training for LVLMs \citep{li2024llavaonevisioneasyvisualtask}, it is likely the case that not all publicly available LVLMs are equal in terms of their ability to utilize ICL examples for classification.

\textbf{Small LVLMs could also become decent emotion reasoners with EmoGist.} As many practical uses of visual emotion understanding often take place within resource-constrained systems with low latency requirements such as web applications or personal computing devices, even relatively small 7 billion models may be beyond typical computing budget under such scenarios. In \autoref{tab:smol}, we test 2 small LVLMs with the same number of 2 billion parameters, SmolVLM2 2.2B \citep{marafioti2025smolvlm} and InternVL2.5 2B-MPO \citep{wang2024mpo}, to examine whether EmoGist performance benefits hold for these smaller models.

We could see that $\text{EmoGist}_{\text{e}}$ achieve similar levels of performance gains over the Zero-Shot and ICL baselines, with SmolVLM2.2 achieving performances similar to the 7B models in \autoref{tab:results}. Given that EmoGist only requires storing cluster centroids and text descriptions at test time, we believe that EmoGist shows some interesting future directions for implementing visual emotion understanding into a wide variety of applications.

\begin{table}[h]
    \centering
    \begin{adjustbox}{width=\columnwidth}
    \begin{tabular}{cc|ccc|ccc}
        \toprule
        \multirow{2}{*}{\textbf{Model}}        & \multirow{2}{*}{\textbf{Method}}  & \multicolumn{3}{c}{\textbf{FI}}    & \multicolumn{3}{c}{\textbf{Memotion}} \\    
                                             &                    &  \msa{Precision} & \msa{Recall} & \msa{F1} &  \msa{Precision} & \msa{Recall} & \msa{F1}  \\
        \midrule
        \multirow{6}{*}{\centering InternVL2.5 2B-MPO}  & \multirow{1.5}{*}{Zero-Shot} \vspace{-0.5em} & 38.039 & 34.813 & 30.499 & 74.272 & 54.972 & 63.181 \\
                                                   & & \varcell{0.285} \vspace{-0.5em} & \varcell{0.320} & \varcell{0.318} & \varcell{0.164} & \varcell{0.129} & \varcell{0.113} \\
                                                   \cmidrule{2-8}
                                                   & \multirow{2.25}{*}{\centering $\text{ICL}_{\text{sim}}$} \vspace{-0.5em} & 45.376 & 11.215 & 16.078 & 64.674 & 75.157 & 69.522 \\
                                                   & & \varcell{0.587} \vspace{-0.5em} & \varcell{0.221} & \varcell{0.309} & \varcell{0.104} & \varcell{0.090} & \varcell{0.074} \\
                                                   &              & \mup{+7.337} & \mdo{-23.598} & \mdo{-14.421} & \mdo{-9.599} & \mup{+20.184} & \mup{+6.341} \\
                                                   \cmidrule{2-8}
                                                   & \multirow{2.25}{*}{\centering $\text{EmoGist}_{\text{e}}$} \vspace{-0.5em} & 50.706 & 41.421 & 37.442 & 70.873 & 60.658 & 65.353 \\
                                                   & & \varcell{0.412} \vspace{-0.5em} & \varcell{0.442} & \varcell{0.675} & \varcell{0.142} & \varcell{0.840} & \varcell{0.504} \\
                                                   &              & \mup{+12.667} & \mup{+6.608} & \mup{+6.943} & \mdo{-3.400} & \mup{+5.685} & \mup{+2.173} \\
        \midrule
        \multirow{6}{*}{\centering SmolVLM2 2.2B} & \multirow{1.5}{*}{Zero-Shot} \vspace{-0.5em}  & 39.567 & 30.762 & 20.703 & 73.124 & 67.617 & 70.263 \\
                                                   & & \varcell{0.056} \vspace{-0.5em} & \varcell{0.019} & \varcell{0.016} & \varcell{0.011} & \varcell{0.012} & \varcell{0.010} \\
                                                       \cmidrule{2-8}
                                                       & \multirow{2.25}{*}{\centering $\text{ICL}_{\text{sim}}$} \vspace{-0.5em} & 47.478 & 48.463 & 41.798 & 74.896 & 12.113 & 20.852 \\
                                                   & & \varcell{0.030} \vspace{-0.5em} & \varcell{0.029} & \varcell{0.028} & \varcell{0.097} & \varcell{0.087} & \varcell{0.130} \\
                                                       &              & \mup{+7.912} & \mup{+17.700} & \mup{+21.096} & \mup{+1.773} & \mdo{-55.505} & \mdo{-49.411} \\
                                                       \cmidrule{2-8}
                                                       & \multirow{2.25}{*}{\centering $\text{EmoGist}_{\text{e}}$} \vspace{-0.5em} & 52.358 & 50.641 & 46.713 & 67.455 & 96.311 & 79.329 \\
                                                   & & \varcell{0.542} \vspace{-0.5em} & \varcell{0.098} & \varcell{0.445} & \varcell{0.381} & \varcell{0.540} & \varcell{0.118} \\
                                                       &              & \mup{+12.791} & \mup{+19.879} & \mup{+26.010} & \mdo{-5.668} & \mup{+28.694} & \mup{+9.066} \\
        \bottomrule
    \end{tabular}
    \end{adjustbox}
    \caption{Results on small VLMs with 2B parameters.}
    \label{tab:smol}
\end{table}

\textbf{Knowledge transfer across different domains with EmoGist.} Lastly, we explore whether the knowledge about different emotion labels we acquire from example images could be used for predictions against the images from the domains different from example images. Using the label descriptions obtained from the example images of the FI dataset, we evaluate $\text{EmoGist}_{\text{e}}$ on the ArtPhoto dataset \citep{10.1145/1873951.1873965}, a collection of artistically photographed images annotated with the same class labels as FI.

In \autoref{tab:cross}, we can see that while EmoGist achieves slightly better scores over naive ICL, the overall performance is actually worse than the Zero-Shot baselines. As EmoGist captures more context-specific, nuanced knowledge of emotion labels, there seems to be a significant semantic gap between the images from FI and ArtPhoto that most of the FI clusters do not adequately explain the test images from ArtPhoto. Making EmoGist to capture both general and context-specific knowledge of emotions across different subject domains and visual compositions could be an interesting direction for future research.

\begin{table}[h]
    \centering
    \begin{adjustbox}{width=\columnwidth}
    \begin{tabular}{cc|ccc}%|ccc}
        \toprule
        \multirow{2}{*}{\textbf{Model}}        & \multirow{2}{*}{\textbf{Method}} & \multicolumn{3}{c}{\textbf{ArtPhoto}} \\    
                                             &                    &  \msa{Precision} & \msa{Recall} & \msa{F1}  \\
        \midrule
        \multirow{3.5}{*}{\centering Qwen2.5-VL 7B} & Zero-Shot    & 52.594 & 42.306 & 41.178  \\
                                                  \cmidrule{2-5}
                                                  & $\text{ICL}_{\text{sim}}$ & 43.332 & 33.165 & 33.745  \\
                                                  %&              & \mdo{-1.64\%} & \mup{0.11\%} & \mup{0.39\%}  \\
                                                  \cmidrule{2-5}
                                                  & $\text{EmoGist}_{\text{e}}$ & 46.535 & 38.555 & 39.227  \\
                                                  %&              & \mdo{-1.64\%} & \mup{0.11\%} & \mup{0.39\%} \\
        \midrule
        \multirow{3.5}{*}{\centering Aya Vision 8B}  & Zero-Shot    & 55.602 & 43.706 & 43.473  \\
                                                   \cmidrule{2-5}
                                                   & $\text{ICL}_{\text{sim}}$ & 43.328 & 26.902 & 26.742    \\
                                                   %&              & \mdo{-1.64\%} & \mup{0.11\%} & \mup{0.39\%} \\
                                                   \cmidrule{2-5}
                                                   & $\text{EmoGist}_{\text{e}}$ & 46.216 & 38.307 & 38.921  \\
                                                   %&              & \mdo{-1.64\%} & \mup{0.11\%} & \mup{0.39\%} \\
        \midrule
        \multirow{3.5}{*}{\centering InternVL2.5 8B-MPO} & Zero-Shot  & 48.824 & 42.365 & 43.518  \\
                                                       \cmidrule{2-5}
                                                       & $\text{ICL}_{\text{sim}}$ & 39.606 & 32.083 & 33.018   \\
                                                       %&              & \mdo{-1.64\%} & \mup{0.11\%} & \mup{0.39\%} \\
                                                       \cmidrule{2-5}
                                                       &  $\text{EmoGist}_{\text{e}}$ & 47.739 & 39.104 & 39.993  \\
                                                       %&              & \mdo{-1.64\%} & \mup{0.11\%} & \mup{0.39\%} \\
        \bottomrule
    \end{tabular}
    \end{adjustbox}
    \caption{Results on ArtPhoto with the emotion label descriptions from FI.}
    \label{tab:cross}
\end{table}

\section{Conclusion and Future Work}
\label{sec:conclusion}

In this paper, we introduced EmoGist, a training-free in-cotext learning method for visual emotion understanding with large vision-language models. We observe a significant amount of improvements over the zero-shot and naive ICL baselines across SOTA LVLMs of 2 and 7 billion parameters. In particular, we find that $\text{EmoGist}_{\text{e}}$, the variant of our method with simple ensembling, achieves robust performance improvements with higher precision. In future work, we'd like to explore more deeply into the label descriptions generated by the strong LVLMs and investigate various reasoning strategies for obtaining emotion label descriptions that are more transferrable across different domains.

\section*{Acknowledgments}
The work
was supported by NSF CAREER award IIS2048001 and the DARPA CCU program. Contents do not necessarily represent the official views of, nor an endorsement by, DARPA, or the US Government.

\section*{Limitations}

Due to limited computational resources available to the authors, we perform our experiments on a limited subset of publicly available large vision-language models with 2 and 7 billion parameters. While we anticipate that our findings would hold overall for other model sizes, we do not provide any direct evidence. We also note that we only tried one model each for the embedding model and the description generation LVLM, as stated in \autoref{sec:emogist}.

%\section*{Acknowledgments}
%
%This document has been adapted by Steven Bethard, Ryan Cotterell and Rui Yan.


\bibliography{custom}

\newpage
\clearpage
\appendix

\section{Dataset Information}
\label{sec:dataset}

For each dataset, we treat the training set as the pool of example images for description generation, use the validation set for hyperparameter tuning, and test our method and baselines against the test set.

\subsection{Memotion}

We use the split introduced by \citet{jin-etal-2024-mm} for our experiments.\footnote{\url{https://huggingface.co/datasets/Ahren09/MMSoc_Memotion}}. Dataset statistics are provided below:

\begin{table}[H]
    \centering
    \begin{adjustbox}{width=\columnwidth}
    \begin{tabular}{c|ccccc}
        \toprule
        \textbf{Split} & \textbf{Number of unique images} & \textbf{sarcastic} & \textbf{humorous} & \textbf{offensive} & \textbf{motivational} \\
        \midrule
        Training & 5593 & 4367 & 4259 & 3417 & 1972 \\
        Validation & 699 & 538 & 539 & 437 & 253 \\
        Test & 700 & 543 & 543 & 425 & 242 \\
        \bottomrule
    \end{tabular}
    \end{adjustbox}
    \caption{Memotion dataset statistics.}
    \label{tab:dataset-memotion}
\end{table}

\subsection{FI}

For the test set, we use the FI hard set introduced by \citep{bhattacharyya-wang-2025-evaluating}. As there is no separate training and validation set provided, we filter all the images includes in the FI hard set from the original FI set, and randomly split the set of remaining images into the training and validation set. Dataset statistics are provided below:

\begin{table}[H]
    \centering
    \begin{adjustbox}{width=\columnwidth}
    \begin{tabular}{c|cccccccc}
        \toprule
        \textbf{Split} & \textbf{amusement} & \textbf{anger} & \textbf{awe} & \textbf{contentment} & \textbf{disgust} & \textbf{excitement} & \textbf{fear} & \textbf{sadness} \\
        \midrule
        Training & 3608 & 842 & 2698 & 4309 & 1391 & 2592 & 852 & 2634  \\
        Validation & 190 & 45 & 142 & 259 & 74 & 137 & 45 & 139  \\
        Test & 1125 & 368 & 293 & 188 & 192 & 185 & 149 & 128 \\
        \bottomrule
    \end{tabular}
    \end{adjustbox}
    \caption{FI dataset statistics.}
    \label{tab:dataset-fi}
\end{table}

\section{Experimental Settings}
\label{sec:test-settings}

\subsection{Hardwares and Softwares Used}

To fully utilize all the GPU resources available to us, we ran all of our validation and test predictions, and embedding generations using multiple NVIDIA V100, GTX TITAN Xp and GTX TITAN X GPUs. 

We use the version 4.50.0 of HuggingFace Transformers library \citep{wolf2020huggingfaces}, the version 0.7.3 of vLLM \citep{kwon2023efficient} alongside PyTorch version 2.5.1 \citep{NEURIPS2019_9015}.

For our label description generation, we used a Mac Studio hardware with M1 Ultra CPU and 128GB of RAM, using the version 0.25.0 of mlx \citep{mlx2023} and 0.1.23 of mlx-vlm.

\subsection{Random Seeds}

For the results shown in \autoref{tab:results} and \autoref{tab:smol}, we run each test for 6 different random seeds: \texttt{21, 42, 63, 84, 105, 126}.

\section{Sensitivity analysis and hyperparameter tuning for the number of clusters $k$}
\label{sec:hyperparameter_tuning}

We determine the optimal number of clusters for each dataset and test-time LVLM combination. Initially, we apply $k$-means clustering to both FI and Memotion datasets for $k$ values of 2, 4, and 6. Subsequently, we generate cluster-based label descriptions for each $k$ using the Qwen2.5-VL 72B model. We then classify the validation set based on these clusters and text descriptions, 3 times for each 7B/2B LVLM using the following random seeds: 21, 42, 63. The optimal $k$ is ultimately identified by identifying $k$ that achieved the most number of highest validation F1 scores across the three seeds. If there's no winner, we ran the validation for additional seeds (84, 105, 126) until the clear winner is found.

\begin{figure*}[h]
\centering
\includegraphics[width=\linewidth]{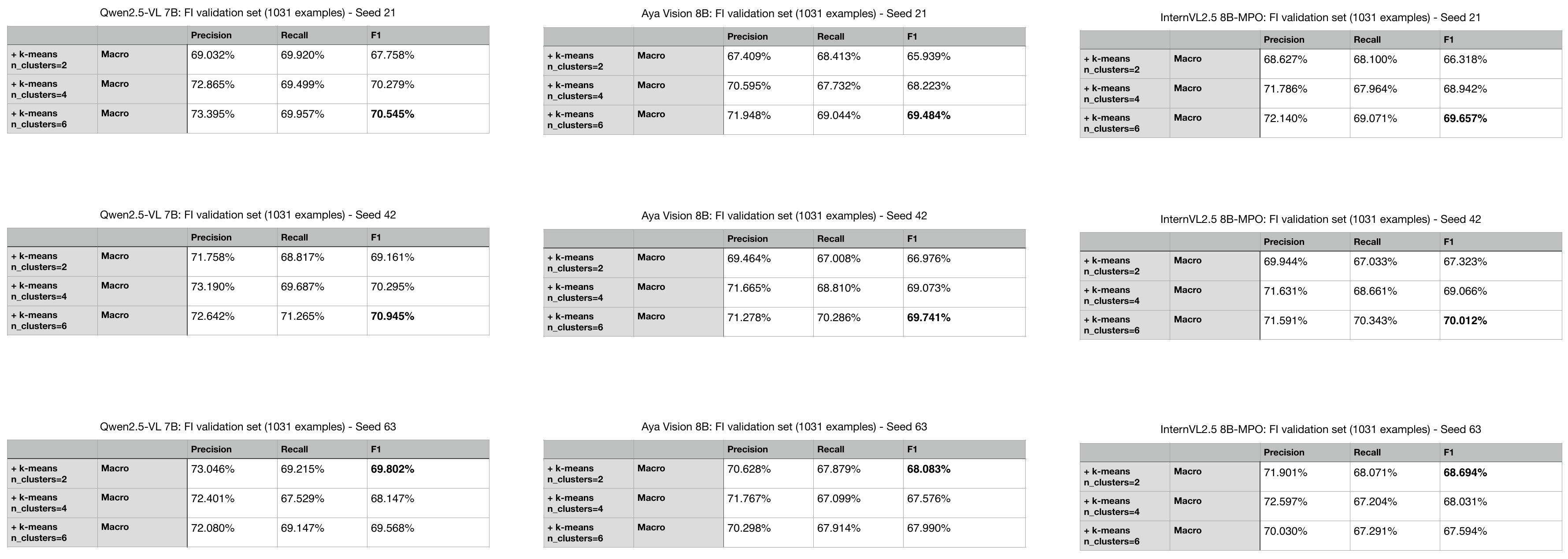}
\caption{Validation set results for $\text{EmoGist}_{\text{n}}$ with 7B models on FI.}
\label{fig:sensitivity_emogist_n_fi}
\end{figure*}

\begin{figure*}[h]
\centering
\includegraphics[width=\linewidth]{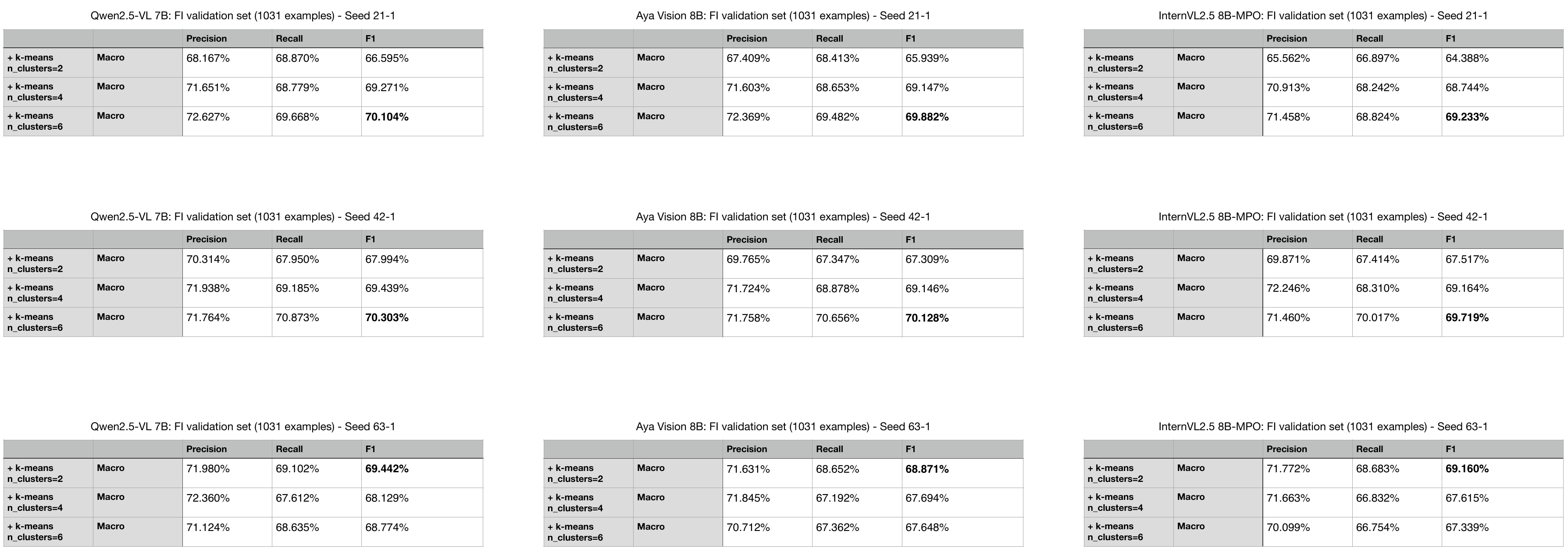}
\caption{Validation set results for $\text{EmoGist}_{\text{e}}$ with 7B models on FI.}
\label{fig:sensitivity_emogist_e_fi}
\end{figure*}

\begin{figure*}[h]
\centering
\includegraphics[width=\linewidth]{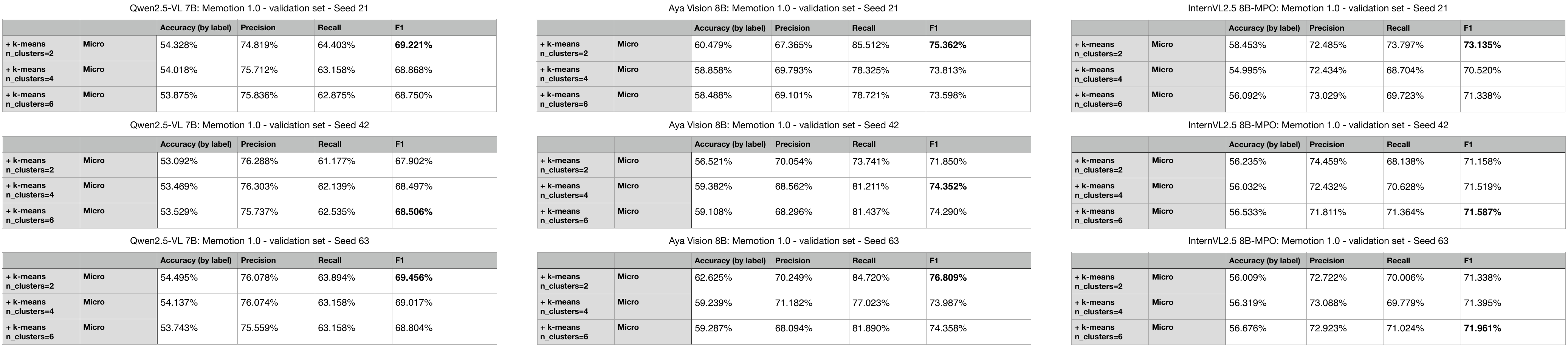}
\caption{Validation set results for $\text{EmoGist}_{\text{n}}$ with 7B models on Memotion.}
\label{fig:sensitivity_emogist_n_memotion}
\end{figure*}

\begin{figure*}[h]
\centering
\includegraphics[width=\linewidth]{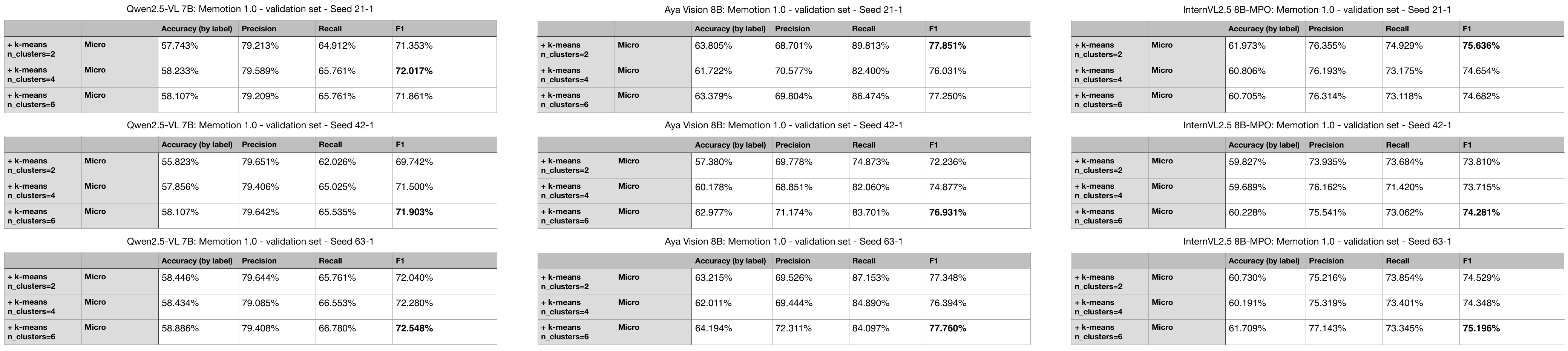}
\caption{Validation set results for $\text{EmoGist}_{\text{e}}$ with 7B models on Memotion.}
\label{fig:sensitivity_emogist_e_memotion}
\end{figure*}

\begin{figure*}[h]
\centering
\includegraphics[width=\linewidth]{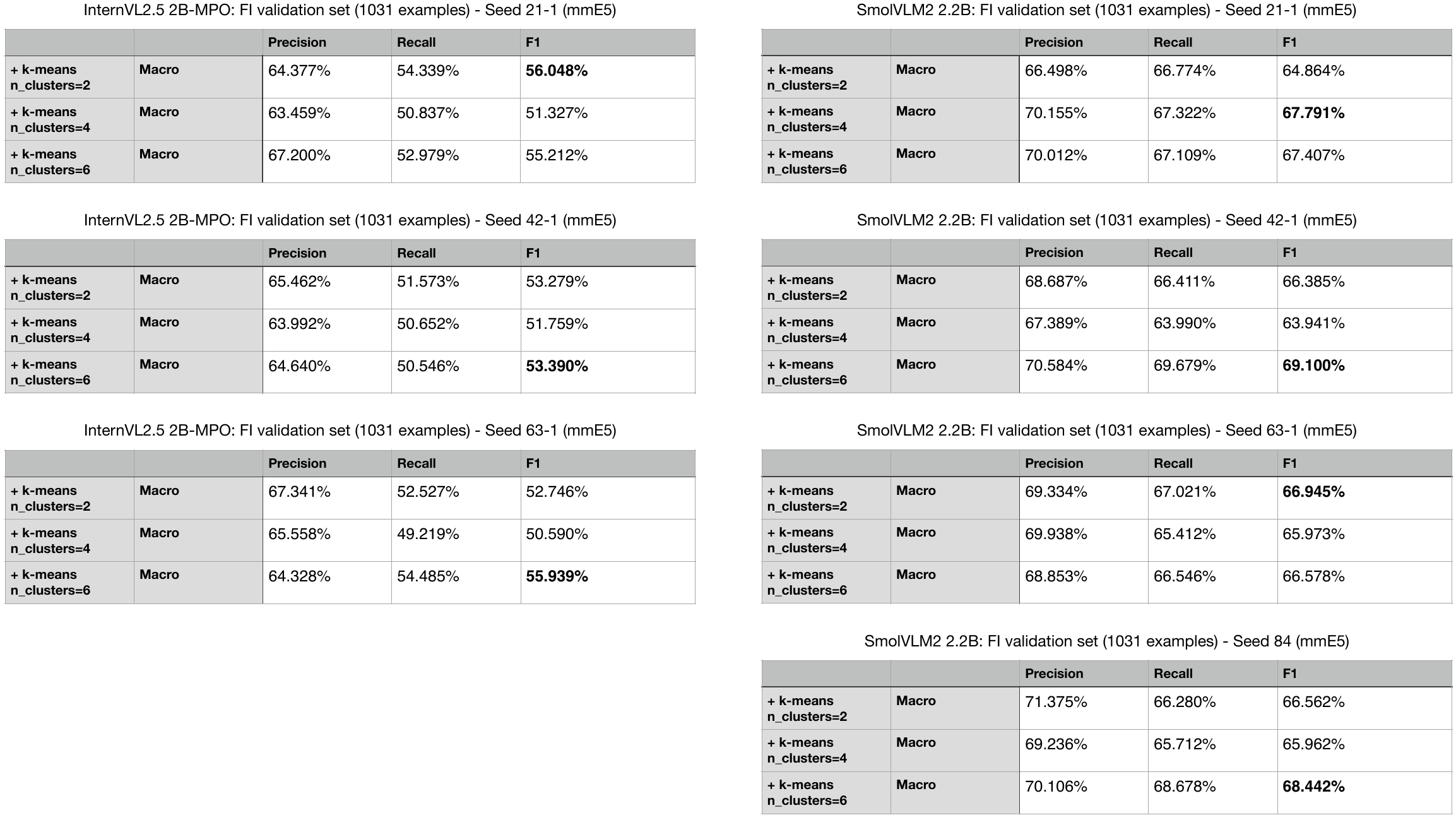}
\caption{Validation set results for $\text{EmoGist}_{\text{e}}$ with 2B models on FI.}
\label{fig:sensitivity_emogist_e_fi_2b}
\end{figure*}

\begin{figure*}[h]
\centering
\includegraphics[width=\linewidth]{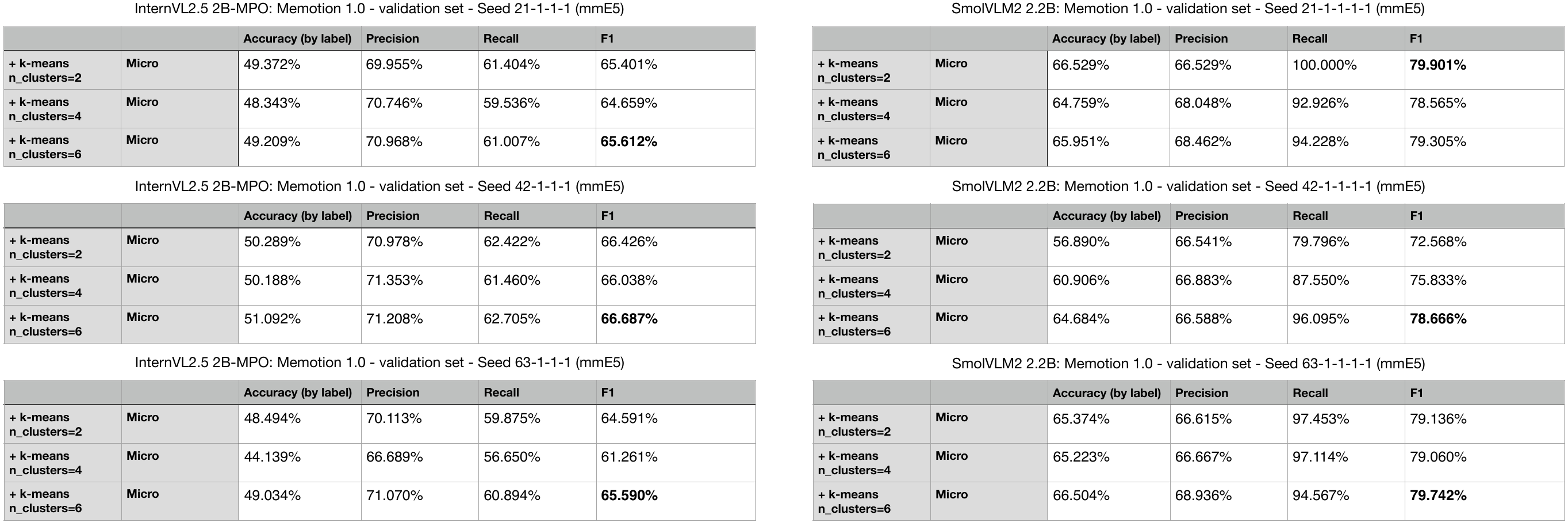}
\caption{Validation set results for $\text{EmoGist}_{\text{e}}$ with 2B models on Memotion.}
\label{fig:sensitivity_emogist_e_memotion_2b}
\end{figure*}

Please see \autoref{fig:sensitivity_emogist_n_fi} and \autoref{fig:sensitivity_emogist_e_fi} for the validation set results of $\text{EmoGist}_{\text{n}}$ and $\text{EmoGist}_{\text{e}}$ with 7B models on FI, \autoref{fig:sensitivity_emogist_n_memotion} and \autoref{fig:sensitivity_emogist_e_memotion} for the validation set results of $\text{EmoGist}_{\text{n}}$ and $\text{EmoGist}_{\text{e}}$ with 7B models on Memotion, and \autoref{fig:sensitivity_emogist_e_fi_2b} and \autoref{fig:sensitivity_emogist_e_memotion_2b} for the validation set results of $\text{EmoGist}_{\text{e}}$ with 2B models on FI and Memotion.

\setlength{\floatsep}{0pt}

\section{Prompts used for label description generation}
\label{sec:prompt_label_description}

\begin{figure}[H]
\centering
\fontsize{7}{9}\selectfont
\begin{tabularx}{\columnwidth}{X}
\\
\hline
\texttt{\color{cyan}**Images from the cluster go here**}\\
These are the examples of images with the sentiment of ``\texttt{\{s\_label\}}". Based on these examples, what are the common features of images to be felt ``\texttt{\{s\_label\}}" by its viewers? Do not reference example images directly.\\
\hline
\end{tabularx}
%\vspace{-1em}
\caption{Prompts used for label description generation.}
\label{fig:prompt_label_description}
\end{figure}

\section{Prompts used for zero-shot and EmoGist}
\label{sec:prompt_classification}

\begin{figure}[H]
\centering
\fontsize{7}{9}\selectfont
\begin{tabularx}{\columnwidth}{X}
\\
\hline
\texttt{\color{cyan}**Test image goes here**} \\
\texttt{\color{orange}**For EmoGist, label description goes here**} \\
Question: Does this image match the sentiment label `\texttt{\{s\_label\}}'? Answer: \\
\\
Answer with `Yes' or `No'.\\
\hline
\end{tabularx}
%\vspace{-1em}
\caption{Prompts used for multi-label zero-shot classification and EmoGist.}
\label{fig:prompt_multilabel_zero_shot}
\end{figure}

\begin{figure}[H]
%\vspace{-5em}
\centering
\fontsize{7}{9}\selectfont
\begin{tabularx}{\columnwidth}{X}
\\
\hline
\texttt{\color{cyan}**Test image goes here**} \\
\texttt{\color{orange}**For EmoGist, label description goes here**} \\
Question: Which of the sentiment labels in the following list does this image belong to? List: [``amusement", ``anger", ``awe", ``contentment",  ``disgust", ``excitement", ``fear", ``sadness"] Answer: \\
\\
Answer with the exact sentiment label as it appears in the list.\\
\hline
\end{tabularx}
%\vspace{-1em}
\caption{Prompts used for multi-class zero-shot classification and EmoGist.}
\label{fig:prompt_multiclass_zero_shot}
\end{figure}

\section{Prompts used for ICL baselines}
\label{sec:prompt_icl}

\begin{figure}[htbp]
\centering
%\vspace{-3em}
\fontsize{7}{9}\selectfont
\begin{tabularx}{\columnwidth}{X}
\\
\hline
\texttt{\color{brown}**ICL example images 1 to 4 goes here**} \\
\texttt{\color{cyan}**Test image goes here**} \\\\

\texttt{\color{brown}\#\# Image 1} \\
\texttt{\color{brown}Question: Does this image matches the sentiment label '\{s\_label\}'? Answer: Yes} \\\\

\texttt{\color{brown}\#\# Image 2} \\
\texttt{\color{brown}Question: Does this image matches the sentiment label '\{s\_label\}'? Answer: Yes} \\\\

\texttt{\color{brown}\#\# Image 3} \\
\texttt{\color{brown}Question: Does this image matches the sentiment label '\{s\_label\}'? Answer: Yes} \\\\

\texttt{\color{brown}\#\# Image 4} \\
\texttt{\color{brown}Question: Does this image matches the sentiment label '\{s\_label\}'? Answer: Yes} \\\\

\texttt{\#\# Image 5} \\
\texttt{Question: Does this image matches the sentiment label '\{s\_label\}'? Answer: } \\\\

Answer with `Yes' or `No'.\\
\hline
\end{tabularx}
%\vspace{-1em}
\caption{Prompts used for multi-label $\text{ICL}_{\text{sim}}$.}
\label{fig:prompt_multilabel_icl_sim}
\end{figure}

\begin{figure}[htbp]
\centering
\fontsize{7}{9}\selectfont
\begin{tabularx}{\columnwidth}{X}
\\
\hline
\texttt{\color{brown}**ICL example images 1 to 4 goes here**} \\
\texttt{\color{cyan}**Test image goes here**} \\\\

\texttt{\color{brown}\#\# Image 1} \\
\texttt{\color{brown}Question: Which of the sentiment labels in the following list does this image belong to? List: [``amusement", ``anger", ``awe", ``contentment",  ``disgust", ``excitement", ``fear", ``sadness"] Answer: \{s\_label\}} \\\\

\texttt{\color{brown}\#\# Image 2} \\
\texttt{\color{brown}Question: Which of the sentiment labels in the following list does this image belong to? List: [``amusement", ``anger", ``awe", ``contentment",  ``disgust", ``excitement", ``fear", ``sadness"] Answer: \{s\_label\}} \\\\

\texttt{\color{brown}\#\# Image 3} \\
\texttt{\color{brown}Question: Which of the sentiment labels in the following list does this image belong to? List: [``amusement", ``anger", ``awe", ``contentment",  ``disgust", ``excitement", ``fear", ``sadness"] Answer: \{s\_label\}} \\\\

\texttt{\color{brown}\#\# Image 4} \\
\texttt{\color{brown}Question: Which of the sentiment labels in the following list does this image belong to? List: [``amusement", ``anger", ``awe", ``contentment",  ``disgust", ``excitement", ``fear", ``sadness"] Answer: \{s\_label\}} \\\\

\texttt{\#\# Image 5} \\
Question: Which of the sentiment labels in the following list does this image belong to? List: [``amusement", ``anger", ``awe", ``contentment",  ``disgust", ``excitement", ``fear", ``sadness"] Answer: \\
\\
Answer with the exact sentiment label as it appears in the list.\\
\hline
\end{tabularx}
%\vspace{-1em}
\caption{Prompts used for multi-class $\text{ICL}_{\text{sim}}$.}
\label{fig:prompt_multiclass_icl_sim}
\end{figure}

\begin{figure}[htbp]
\centering
%\vspace{-1em}
\fontsize{7}{9}\selectfont
\begin{tabularx}{\columnwidth}{X}
\\
\hline
\texttt{\color{brown}**ICL example images 1 to 8 goes here**} \\
\texttt{\color{cyan}**Test image goes here**} \\\\

\texttt{\color{brown}\#\# Image 1} \\
\texttt{\color{brown}Question: Which of the sentiment labels in the following list does this image belong to? List: [``amusement", ``anger", ``awe", ``contentment",  ``disgust", ``excitement", ``fear", ``sadness"] Answer: amusement} \\\\

\texttt{\color{brown}\#\# Image 2} \\
\texttt{\color{brown}Question: Which of the sentiment labels in the following list does this image belong to? List: [``amusement", ``anger", ``awe", ``contentment",  ``disgust", ``excitement", ``fear", ``sadness"] Answer: anger} \\\\

\texttt{\color{brown}\#\# Image 3} \\
\texttt{\color{brown}Question: Which of the sentiment labels in the following list does this image belong to? List: [``amusement", ``anger", ``awe", ``contentment",  ``disgust", ``excitement", ``fear", ``sadness"] Answer: awe} \\\\

\texttt{\color{brown}\#\# Image 4} \\
\texttt{\color{brown}Question: Which of the sentiment labels in the following list does this image belong to? List: [``amusement", ``anger", ``awe", ``contentment",  ``disgust", ``excitement", ``fear", ``sadness"] Answer: contentment} \\\\

\texttt{\color{brown}\#\# Image 5} \\
\texttt{\color{brown}Question: Which of the sentiment labels in the following list does this image belong to? List: [``amusement", ``anger", ``awe", ``contentment",  ``disgust", ``excitement", ``fear", ``sadness"] Answer: disgust} \\\\

\texttt{\color{brown}\#\# Image 6} \\
\texttt{\color{brown}Question: Which of the sentiment labels in the following list does this image belong to? List: [``amusement", ``anger", ``awe", ``contentment",  ``disgust", ``excitement", ``fear", ``sadness"] Answer: excitement} \\\\

\texttt{\color{brown}\#\# Image 7} \\
\texttt{\color{brown}Question: Which of the sentiment labels in the following list does this image belong to? List: [``amusement", ``anger", ``awe", ``contentment",  ``disgust", ``excitement", ``fear", ``sadness"] Answer: fear} \\\\

\texttt{\color{brown}\#\# Image 8} \\
\texttt{\color{brown}Question: Which of the sentiment labels in the following list does this image belong to? List: [``amusement", ``anger", ``awe", ``contentment",  ``disgust", ``excitement", ``fear", ``sadness"] Answer: sadness} \\\\

\texttt{\#\# Image 9} \\
Question: Which of the sentiment labels in the following list does this image belong to? List: [``amusement", ``anger", ``awe", ``contentment",  ``disgust", ``excitement", ``fear", ``sadness"] Answer: \\
\\
Answer with the exact sentiment label as it appears in the list.\\
\hline
\end{tabularx}
%\vspace{-1em}
\caption{Prompts used for multi-class, $\text{ICL}_{\text{all}}$.}
\label{fig:prompt_multiclass_icl_all}
\end{figure}

\end{document}